\documentclass[runningheads]{llncs}
\usepackage{graphicx}
\usepackage{amsmath,lipsum}
\usepackage{amsfonts}
\usepackage{array}
\usepackage{multirow}
\usepackage{subfigure}
\usepackage{caption}

\newcolumntype{P}[1]{>{\centering\arraybackslash}p{#1}}
\newcolumntype{M}[1]{>{\centering\arraybackslash}m{#1}}

\newcommand{\repeatthanks}{\textsuperscript{\thefootnote}}

\begin{document}
\title{Deep Knowledge Tracing with Side Information}


\author{Zhiwei Wang\thanks{Work was done when the authors did internship in TAL AI Lab}\inst{1} \and
Xiaoqin Feng\repeatthanks\inst{2} \and 
Jiliang Tang\inst{1} \and Gale Yan Huang\inst{2} \and Zitao Liu\thanks{Corresponding Author: Zitao Liu}\inst{2}}

\authorrunning{Z. Wang et al.}

\institute{Data Science and Engineering Lab, Michigan State University, USA  \\
\email{\{wangzh65,tangjili\}@msu.edu}
\and TAL AI Lab, Beijing, China\\
\email{\{fengxqin,galehuang,liuzitao\}@100tal.com}}
\maketitle             

\vspace{-0.4cm}
\begin{abstract}
Monitoring student knowledge states or skill acquisition levels known as knowledge tracing, is a fundamental part of intelligent tutoring systems. Despite its inherent challenges, recent deep neural networks based knowledge tracing models have achieved great success, which is largely from models' ability to learn sequential dependencies of questions in student exercise data. However, in addition to sequential information, questions inherently exhibit side relations, which can enrich our understandings about student knowledge states and has great potentials to advance knowledge tracing. Thus, in this paper, we exploit side relations to improve knowledge tracing and design a novel framework DTKS. The experimental results on real education data validate the effectiveness of the proposed framework and demonstrate the importance of side information in knowledge tracing.
\end{abstract}

\section{Introduction}


Knowledge tracing - where machine monitors students' knowledge states and their skill acquisition levels - is essential for personalized education and a fundamental part of intelligent tutoring systems~\cite{yudelson2013individualized,corbett1994knowledge,d2008more,piech2015deep}. 
However, tracing student knowledge states is inherently challenging because of the complexity of human learning process, which involves a variety of factors from diverse domains such as neural science~\cite{bassett2011dynamic,caine1990understanding}, psychology~\cite{hilgard1948theories}, and education~\cite{felder1988learning}. Meanwhile, the large amount of data produced by a growing number of online education platforms and recent advances of machine learning technology provide us with unprecedented opportunities to build advanced models for accurate knowledge tracing. Consequently, it has garnered widespread attention from researchers in both education and artificial intelligence communities~\cite{piech2015deep,zhang2017dynamic,wang2017deep}. Recently, one framework named Deep Knowledge Tracing (DKT) that is based on deep neural networks has shown superior  performance over previously proposed knowledge tracing models~\cite{piech2015deep}. Specifically, based on student historical answered questions, it is able to predict student performance on future questions with high accuracy. The key reason of the success of DKT is its ability to capture the sequential dependencies among questions embedded in the question answer sequences.

In fact, in addition to the sequential dependencies, questions naturally exhibit side relations due to their intrinsic properties. For example, questions are typically designed to improve certain concepts or skills. Thus, questions with similar underlying concepts or skills are inherently related. These relations can be represented as a question-question graph where nodes are questions and an edge exists in two questions if they are designed to examine similar sets of skills and concepts. The question-question graph provides rich information that can lead us to a better understanding of student knowledge states and exploiting such information has the great potential to improve the knowledge tacking performance. 

In this work, we exploit question relation information for better knowledge tracing and propose a framework DTKS that can capture both sequential dependencies and intrinsic relations of questions simultaneously. In summary, the contributions of this work are: 1) We identify the importance to incorporate side relations of questions into knowledge tracing; 2) We design a novel framework DKTS that provides a principled approach to capture both sequential and side relation information to model the student knowledge states and accurately predict their performance; and 3) We demonstrate the effectiveness of the proposed framework with real data. 
%



\section{Related Work}

In this section, we briefly review the related works. Knowledge tracing is a long established research question and an essential task for computer assisted education~\cite{yudelson2013individualized,d2011ensembling,corbett1994knowledge,d2008more,piech2015deep,zhang2017dynamic}.
Previously, Bayesian Knowledge Tracing (BKT) based approach has been in predominate use~\cite{yudelson2013individualized,d2008more}. It represents the student knowledge state with a set of binary variables and each variable corresponds  to student understanding of a single concept~\cite{corbett1994knowledge}. Other approaches such as Learning Factors Analysis~\cite{cen2006learning} and ensemble methods~\cite{d2011ensembling} have also been proposed and achieved comparable performance with BKT. Recently, deep neural network based approach has become increasingly popular~\cite{piech2015deep,zhang2017dynamic}. Models in this line such as DKT~\cite{piech2015deep} represent student knowledge state with continuous and expressive latent vectors and are able to capture the complexity of knowledge state. However, few of them incorporates the question relation information, which could be very helpful for knowledge tracing tasks.


\section{The Proposed Framework}
\begin{figure}
\centering
\includegraphics[scale=0.3]{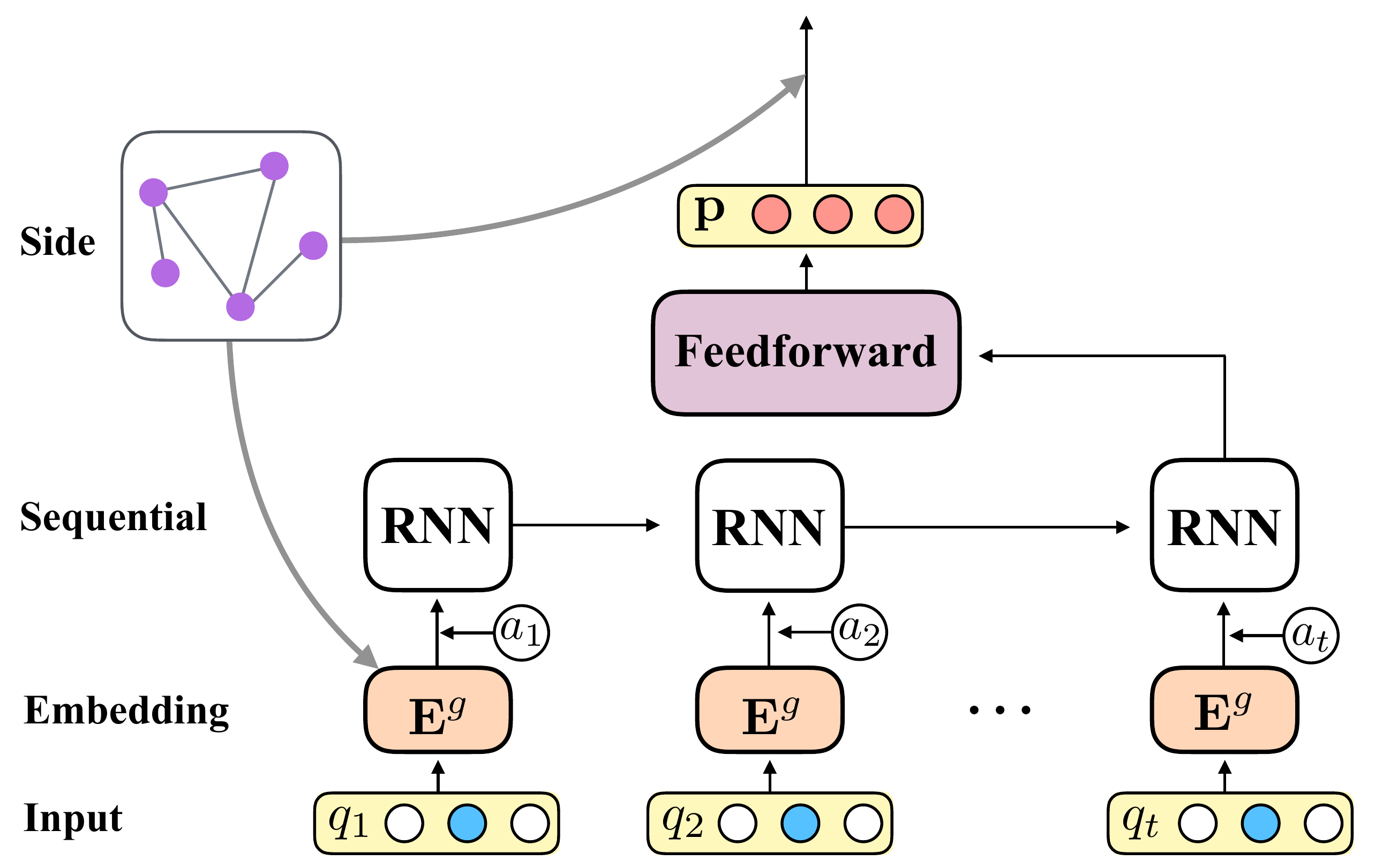}
	\caption{The network architecture of the proposed framework.}
	\label{fig:net}
\end{figure}
In this section, we introduce our proposed model DKTS that is able to incorporate question relations in modeling student knowledge state. The overall structure of the proposed model is shown in Figure~(\ref{fig:net}). Before detailing each layer next, we first introduce the notations. Vectors and matrices are represented with bold lower-case letters such as ${\bf h}$ and  bold upper-case letters such as ${\bf W}$. In addition, the $i^{th}$ entry of vector ${\bf h}$ is denoted as ${\bf h}(i)$ and the entry at the $i^{th}$ row and $j^{th}$ column of matrix ${\bf W}$ as ${\bf W}(i,j)$.

\noindent {\bf Input and Embedding Layers:} The input of the framework is the student past question answer sequence $S=(x_1, x_2, \cdots, x_n)$, where $x_j = (q_j, a_j)$ involves a question $q_j$ and the correctness of the student answer denoted as $a_j \in \{0, 1\}$. We represent $x_j$ as ${\bf x}_j$ using an embedding layer.

\noindent {\bf The Sequential Layer:} We take advantage of RNN models to trace student knowledge states. Specifically, at time step $t$, RNN maintains a latent vector ${\bf h}_t \in \mathbb{R}^{n_h}$ representing student knowledge state through the following cell structure: ${\bf h}_t = tanh({\bf W}{\bf x}_t + {\bf U}{\bf h}_{t-1} + {\bf b})$.  Thus, this recursive structure naturally describes the evolution of student knowledge state ${\bf h}_t$ that is driven by the previous knowledge state ${\bf h}_{t-1}$ and current observation ${\bf x}_t$. In practice, more advanced recurrent cells such as long short-term memory unit (LSTM) and gated recurrent unit cell (GRU)~\cite{hochreiter1997long,cho2014learning} often achieve better performance than original cell. We investigate both of them in this work. After sequential layer, we design a feedforward layer to predict the student future's response to each question based on the final knowledge state representation ${\bf h}$ by following equation:  ${\bf p} = \sigma({\bf h}{\bf W}^p + {\bf b}^p)$, where ${\bf p}(i)$ indicates the probability that the student can answer the $i^{th}$ question correctly. 

\noindent{\bf The Side Layer:} In side layer, two model components are designed to capture the question relation. Firstly, instead of using embedding layers, we apply graph embedding algorithms such as LINE~\cite{tang2015line} and Node2Vec~\cite{grover2016node2vec} to the question-question relation graphs to obtain the question representations that preserve the question relations. Secondly, to impose the intuition that if a pair of questions (e.g., $i^{th}$ and $j^{th}$ questions) requires similar skills or involves similar concepts, the probability for a given student answering the two questions correctly should also be similar, we design the following regularization term $\mathcal{L}_r  = \frac{1}{2} {\bf p}^T {\bf L} {\bf p}$, where ${\bf L}$ is the Laplacian matrix of adjacent matrix ${\bf A}$ representing the question relation graph.

\noindent{\bf The Loss Function: } With the prediction ${\bf p}$ obtained from sequential layer and the relation regularization term $\mathcal{L}_r$, we define the loss function of the proposed framework DKTS for each training data as $\mathcal{L} = \mathcal{L}_p + \alpha \mathcal{L}_r$, where $\alpha$ is adopted to control the contribution of relation regularizer and $\mathcal{L}_p$ is the binary cross-entry loss that is defined as:
\begin{align} 
\mathcal{L}_p =  -a_{t+1} \log ({\bf p}^T{\bf q}_{t+1}) - (1-a_{t+1}) log (1-{\bf p}^T{\bf q}_{t+1})
\end{align}
where ${\bf q}_{t+1}$ is the one-hot encoding of the question at time step $t+1$.

\section{Experiment}
In this section, we conduct experiments on real education data to verify the effectiveness of the proposed model.

\noindent{\bf Dataset:} We collect a student question answer behavior dataset from one of the most popular GMAT preparation mobile applications in China. It contains 8,684 questions and 90831 anonymized students and is cleaned by a filtering process. For each student, we collect her question answer behaviors and form a sequence of behaviors ordering by time information. A question relation graph is constructed according to the underlying knowledge and skills.

\noindent{\bf Baselines:} In baselines, we use RNN, LSTM, and GRU to model the students knowledge state and represent question by embedding vectors that are sampled from a Gaussian distribution $\mathcal{N}({\bf 0}, {\bf I})$ (Gaussian), or obtained through graph embedding algorithms (LINE, Node2Vec). Note that previously proposed DKT model uses LSTM to learn student knowledege state with question representation vectors sampled from Gaussian distribution~\cite{piech2015deep}.

\noindent{\bf Experimental Results:}
We evaluate the prediction performance by area under the curve (AUC) and a higher AUC indicates better performance. The results are shown in Table~{\ref{table:result}}. We observe that 1) The embedding vectors that preserve the question relation information significantly improve the prediction performance, which clearly demonstrates the importance of question relation information for knowledge tracing tasks; and 2) The proposed framework DKTS outperforms all other methods by a large margin. We contribute the superior performance of the proposed model to its ability to incorporate question relation information. 

\begin{table}
    \vspace{-10mm}

	\begin{center}	
	\caption{Performance Comparison Results. `NA' indicates not applicable.}
		\label{table:result}
        \begin{tabular}{| M{2cm} |M{1.8cm} M{1.8cm} M{1.8cm}|}
			\hline 
			\multirow{2}{*}{Method}& \multicolumn{3}{c|}{Question Embedding}  \\
            \cline{2-4}
			 & Gaussian & LINE  &  Node2Vec  \\
			\hline
        RNN & 0.6527  & 0.7015 & 0.6988    \\	
        LSTM & 0.6999 &  0.7152 & 0.7140 \\ 
	    GRU &   0.7074  &  0.7173 & 0.7165   \\ 
	    DKTS & NA & 0.7338 & 0.7340 \\
        \hline
		\end{tabular}
    \vspace{-12mm}
	\end{center}
\end{table}

\section{Conclusion}

In this work, we exploit question relation information for knowledge tracing tasks. Specifically, we design a novel deep neural network based framework that is able to capture the sequential dependencies and intrinsic relations of questions to trace the student knowledge state. Moreover, we evaluate the proposed framework with real education data on student future interaction prediction task. The experimental results have clearly demonstrated the importance of the question relation information and the proposed framework outperforms state-of-the-art baselines significantly.

\vspace{3mm}

\noindent{\bf Acknowledgements.} Zhiwei Wang and Jiliang Tang are supported by the National Science Foundation (NSF) under grant numbers IIS-1714741, IIS-1715940, IIS-1845081 and CNS-1815636, and a grant from Criteo Faculty Research Award.

\bibliographystyle{splncs04}
\bibliography{zhiwei}

\begin{thebibliography}{10}
\providecommand{\url}[1]{\texttt{#1}}
\providecommand{\urlprefix}{URL }
\providecommand{\doi}[1]{https://doi.org/#1}

\bibitem{d2008more}
d~Baker, R.S., Corbett, A.T., Aleven, V.: More accurate student modeling
  through contextual estimation of slip and guess probabilities in bayesian
  knowledge tracing. In: Intelligent Tutoring Systems (2008)

\bibitem{d2011ensembling}
d~Baker, R.S., Pardos, Z.A., Nooraei, B.B., Heffernan, N.T.: Ensembling
  predictions of student knowledge within intelligent tutoring systems. In:
  UMAP (2011)

\bibitem{bassett2011dynamic}
Bassett, D.S., Porter, M.A., Mucha, P.J., Carlson, J.M., Grafton, S.T.: Dynamic
  reconfiguration of human brain networks during learning. PNAS  (2011)

\bibitem{caine1990understanding}
Caine, R.N., Caine, G.: Understanding a brain-based approach to learning and
  teaching. Educational Leadership  \textbf{48}(2),  66--70 (1990)

\bibitem{cen2006learning}
Cen, H., Koedinger, K., Junker, B.: Learning factors analysis--a general method
  for cognitive model evaluation and improvement. In: ITS. Springer (2006)

\bibitem{cho2014learning}
Cho, K., Van~Merri{\"e}nboer, B., Gulcehre, C., Bahdanau, D., Bougares, F.,
  Schwenk, H., Bengio, Y.: Learning phrase representations using rnn
  encoder-decoder for statistical machine translation. arXiv preprint
  arXiv:1406.1078  (2014)

\bibitem{corbett1994knowledge}
Corbett, A.T., Anderson, J.R.: Knowledge tracing: Modeling the acquisition of
  procedural knowledge. User modeling and user-adapted interaction  (1994)

\bibitem{felder1988learning}
Felder, R.M., Silverman, L.K., et~al.: Learning and teaching styles in
  engineering education. Engineering education  \textbf{78}(7),  674--681
  (1988)

\bibitem{grover2016node2vec}
Grover, A., Leskovec, J.: node2vec: Scalable feature learning for networks. In:
  KDD (2016)

\bibitem{hilgard1948theories}
Hilgard, E.R.: Theories of learning.  (1948)

\bibitem{hochreiter1997long}
Hochreiter, S., Schmidhuber, J.: Long short-term memory. Neural computation
  (1997)

\bibitem{piech2015deep}
Piech, C., Bassen, J., Huang, J., Ganguli, S., Sahami, M., Guibas, L.J.,
  Sohl-Dickstein, J.: Deep knowledge tracing. In: NIPS (2015)

\bibitem{tang2015line}
Tang, J., Qu, M., Wang, M., Zhang, M., Yan, J., Mei, Q.: Line: Large-scale
  information network embedding. In: WWW (2015)

\bibitem{wang2017deep}
Wang, L., Sy, A., Liu, L., Piech, C.: Deep knowledge tracing on programming
  exercises. In: L@S (2017)

\bibitem{yudelson2013individualized}
Yudelson, M.V., Koedinger, K.R., Gordon, G.J.: Individualized bayesian
  knowledge tracing models. In: AIED (2013)

\bibitem{zhang2017dynamic}
Zhang, J., Shi, X., King, I., Yeung, D.Y.: Dynamic key-value memory networks
  for knowledge tracing. In: WWW (2017)

\end{thebibliography}
\end{document}